# A Defeasible Deontic Calculus for Resolving Norm Conflicts

**Taylor Olson**  TAYLOROLSON@U.NORTHWESTERN.EDU
**Roberto Salas-Damian**  ROBERTO.SALAS@U.NORTHWESTERN.EDU
**Kenneth D. Forbus**  FORBUS@NORTHWESTERN.EDU
Department of Computer Science, Northwestern University, Evanston, IL 60208 USA

## Abstract

When deciding how to act, we must consider other agents' norms and values. However, our norms are ever-evolving. We often add exceptions or change our minds, and thus norms can conflict over time. Therefore, to maintain an accurate mental model of other's norms, and thus to avoid social friction, such conflicts must be detected and resolved quickly. Formalizing this process has been the focus of various deontic logics and normative multi-agent systems. We aim to bridge the gap between these two fields here. We contribute a defeasible deontic calculus with inheritance and prove that it resolves norm conflicts. Through this analysis, we also reveal a common resolution strategy as a red herring. This paper thus contributes a theoretically justified axiomatization of norm conflict detection and resolution.

## 1. Introduction

Living in a social world, we must be able to govern our own actions with respect to the norms of other agents and institutions. For instance, when determining if you should help your sister cook, you ought to consider her norms, for she may have told you not to intervene. However, the dynamic nature of our social environment often produces conflicts between norms. That is, the same action may seem to be both required and prohibited (or optional). Your sister may have said weeks ago, "Do not help me cook" yet just now said, "Help me cook these vegetables." What should you do? To make this determination, you must be capable of detecting and resolving such conflicts quickly.

In this paper, we present the Defeasible Deontic Inheritance Calculus (DDIC), a formalism for automatically resolving a continuous stream of possibly conflicting norms. We aim to unify recent developments within the normative multi-agent systems (NMAS) community (Santos et al., 2017) with more theoretical work on deontic and defeasible logic (Horty, 1994). We specifically show that the resolution strategies common to NMAS can be explained under deontic inheritance given that we incorporate defeasibility. We thus advance the state of the art in automated norm conflict detection and resolution.

We start by providing background on norms and default and deontic logic. Next, we formalize norm conflict resolution with defeasible deontic inheritance in DDIC. We then prove that DDIC axiomatizes common resolution strategies, further revealing an existing strategy as a red herring. We conclude with a discussion of future research.

## 2. Background

A *norm* (also called a normative belief) is defined here as an agent's evaluative attitude of a behavior in a given context. For instance, "Leo believes cooking in the morning is impermissible." More formally, a norm is a tuple $\langle A, B, C, D \rangle$, where $A$ is an agent holding the belief, $B$ and $C$ are formal representations of a behavior and context, and $D$ is a deontic modal $\in \{Obl, Opt, Imp\}$ (taken as "Obligatory", "Optional", and "Impermissible"). Similar formal representations for norms are used in cognitive science and AI research (Bello & Malle, 2023; Olson & Forbus, 2023), normative multi-agent systems (Santos et al., 2017; Andrighetto et al., 2010; Bringsjord & Sundar, 2013), and conditional deontic logics (van Fraassen, 1972; Castañeda, 1989).

The *grounds of application* of a norm consists of all acts or behaviors in which that norm applies (Elhag et al., 1999). Two norms then conflict at a context when they share application grounds, but their deontic statuses are inconsistent. For instance, the obligation to help your sister and the prohibition against helping her cook, both share the application "helping cook", and thus seemingly conflict. Various norm conflict types thus arise from the intricate relationships between behaviors, contexts, and deontic statuses. Building on ideas from Ross (1958) and those of the multi-agent community, two norms can conflict in three ways (visualized in Figure 1):

**Direct Conflict (Ross's Total-Total):** Two norms directly conflict at context $C$ when their behaviors are equivalent, their contexts intersect at $C$, and their evaluations are inconsistent. I.e., they have identical application grounds at $C$ but make inconsistent evaluative claims.

**Indirect Conflict (Ross's Total-Partial):** Two norms indirectly conflict at context $C$ when the behavior of one norm entails (and is not equivalent to) the behavior of the other, their contexts intersect at $C$, and their evaluations are inconsistent. I.e., the application grounds of one norm subsumes the other at $C$ but they make inconsistent evaluative claims.

**Intersecting Conflict (Ross's Intersection):** Two norms conflict at intersection $S$ in context $C$, when their behaviors intersect at $S$ (and neither behavior subsumes the other), their contexts intersect at $C$, and their evaluations are inconsistent. I.e., their application grounds share only a proper subset and make inconsistent evaluative claims.

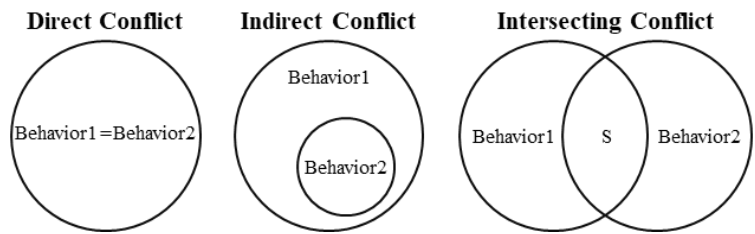

*Figure 1.* Venn Diagrams of Direct, Indirect, and Intersecting Norm Conflicts.

The ability to resolve such conflicts is fundamental for cohesive social interaction. Thus, various strategies have been developed for multi-agent systems. The three heuristics commonly deployed (Santos et al., 2017) are: *Lex Specialis*—prioritize the most specific norm; *Lex Posterior*—prioritize the most recent norm; and *Lex Superior*—prioritize the highest authority. Approaches then iteratively apply such heuristics to yield a conflict free set of norms.

### 2.1 Standard Deontic Logic and Deontic Inheritance

Central to our approach for detecting norm conflicts is the *principle of Inheritance* (Ross, 1941), or *OB-RM*, from standard deontic logic (SDL). We show that this principle can be used to determine if and where norms conflict. Briefly, the semantics of SDL draw upon Kripke-style possible world semantics (Kripke, 1963). SDL grounds deontic modality in the concept of Ideal worlds, captured by an accessibility relation $R$ over possible worlds. Thus, $xRy$ holds that world $y$ is Ideal, relative to world $x$. Given these semantics, $Obl(p)$ holds at world $x$ if and only if $p$ is true in every ideally accessible world $y$; $Imp(p)$ holds at world $x$ if $p$ is false at every such ideal world; And $Opt(p)$ holds if $p$ is neither uniformly true nor false at every such world.

OB-RM can then be derived given SDL's semantics: *if $\vdash p \to q$, then $\vdash Obl(p) \to Obl(q)$*. Intuitively, if whenever $p$ is true, $q$ is also true, then if $p$ is true at all ideally accessible worlds, $q$ must also be true at all ideally accessible worlds. Though we do not ground deontics in possible world semantics here, the idea of deontic inheritance is central to our approach.

### 2.2 Defeasible Reasoning

While we utilize deontic inheritance to detect conflicts, we utilize defeasible reasoning to resolve them. We specifically build upon Reiter's (1980) default logic, which introduces the notion of *default rules*. Unlike standard deductive rules, default rules consider that certain inferences admit to exceptions. Such rules are of the form:

$$\frac{Prerequisite : Justification_{1,\ldots,n}}{Consequent}$$

A default rule states that if the Prerequisite is true, then we can conclude the Consequent so long as each Justification is consistent with our current beliefs. More formally, given a theory $T$, we apply default rule $\frac{P:J_1,\ldots,n}{C}$ on theory $T$ when $T \vdash P$ and for all $J_n, T \nvdash \neg J_n$, yielding $T = T \cup \{C\}$.

A *default theory* is a pair of the form $\langle T, D \rangle$ where $T$ is a set of background logical formulae and $D$ is a set of default rules. To compute conclusions of a theory $T$, rules from $D$ are exhaustively applied to $T$ until no other rules can be applied. This conclusion set is called an *extension* of the theory. Theory $T \vdash C$ when, after applying rules on $T$, $C \in T$. However, because rules may be applied in various orders, a theory can produce multiple (possibly contradictory) extensions (e.g., Nixon diamond cases). To avoid such cases, some formalisms define an explicit ordering amongst default rules (Horty, 2007): $P(R1, R2)$ means that $R1$ is to be applied before $R2$.

## 3. Resolving Conflicts via Defeasible Reasoning

While a norm is defined here as an internal attitude, *normative testimony* can be taken as its external (natural language) expression. Thus, norms are inferred from normative testimony. For instance, from your sister's first claim, "Do not help me cook", we might infer she believes we should not help her cook *vegetables*, *while wearing boots*, *when it's raining*. However, such inferences should be defeated when we infer norms from her later claim, "Help me cook these vegetables." Detecting and resolving such conflicts is vital to maintaining an accurate mental model of other agents' norms. We formalize this process within a defeasible deontic calculus in the sections below.

### 3.1 The Defeasible Deontic Inheritance Calculus (DDIC)

Let $L$ be a standard propositional language for which the norms govern[1]. Take $L$ to contain the usual symbols of $\wedge, \vee, \rightarrow,$ and $\neg$ for the operations of conjunction, disjunction, implication, and negation under usual interpretations. Take $\top$ as the proposition that is trivially true and turnstile $\vdash$ as ordinary logical consequence. We assume a standard transitivity rule amongst implied sentences of $L$: $[a \rightarrow b \rightarrow c] \rightarrow [a \rightarrow c]$. Such relations are at the heart of deontic inheritance.

The language for the Defeasible Deontic Inheritance Calculus (DDIC) is then obtained from $L$ by adding deontic modals for normative beliefs: $Obl$, $Imp$ and $Opt$. Normative testimony is represented with accents: $\ddot{O}bl$, $\ddot{O}pt$, $\dddot{Imp}$. This distinction between an agent's external testimony and their inferred internal belief is critical for computing defeaters. Normative formulae also contain a time parameter, as ordering is important for conflict resolution. We assume this language $Time$ is isomorphic with $\mathbb{N}$ given the relations $>, <, \geq, \leq$ under standard interpretation. We formally define DDIC below.

***Definition 1*** *(Deontic Language DDIC).* If $a$ is a positive literal, $\varphi$ is a formula of $L$, $t \in Time$:

- $Obl(a, \varphi, t), Imp(a, \varphi, t)$ and $Opt(a, \varphi, t)$ are normative belief formulae of $DDIC$ to be read as "at time $t$ it *is believed* that given $\varphi$ is true, $a$ is obligatory", "… impermissible", and "… optional" respectively.
- $\ddot{O}bl(a, \varphi, t), \dddot{Imp}(a, \varphi, t)$ and $\ddot{O}pt(a, \varphi, t)$ are normative testimony formulae of $DDIC$ to be read as "at time $t$ it *was said* that given $\varphi$ is true, $a$ is obligatory", "… impermissible", and "… optional" respectively.
- If $f$ is a normative formula of $DDIC$, then $\neg f$ is a normative formula of $DDIC$.

A few notes on this language. First, normative formulae of $DDIC$ are true with respect to an agent, or normative system, but to simplify notation we leave the agent implicit. I.e., $Imp(a, \varphi, t)$ can truly be taken as stating that "at time $t$, *this agent* believes that given $\varphi$ is true, $a$ is impermissible." Second, though we limit a norm's behavior to positive literals, we can represent the deontic statuses of any intended negated behavior as well in a corresponding positive form: $Obl(\neg a) \stackrel{\text{def}}{=} Imp(a)$. Lastly, we call normative formulae of the form $\ddot{O}bl(a, \varphi, t)$ *"obligations"*, $\dddot{Imp}(a, \varphi, t)$ *"prohibitions"*, and $\ddot{O}pt(a, \varphi, t)$ *"discretionary norms"*.

*3.1.1 Axioms of DDIC*

The following axioms formalize defeasible deontic inheritance in DDIC. For each rule below, take $\varphi, \gamma,$ and $\delta$ as formulae of $L$ and lowercase letters (e.g., $a, b, z, y$) as positive literals of $L$.

D1a. $\dfrac{}{\ddot{O}pt(a, \varphi, t) \leftrightarrow [\neg \ddot{O}bl(a, \varphi, t) \wedge \neg \dddot{Imp}(a, \varphi, t)]}$   D1b. $\dfrac{}{\ddot{O}bl(a, \varphi, t) \rightarrow \neg \dddot{Imp}(a, \varphi, t)}$

D1c. $\dfrac{}{Opt(a, \varphi, t) \leftrightarrow [\neg Obl(a, \varphi, t) \wedge \neg Imp(a, \varphi, t)]}$   D1d. $\dfrac{}{Obl(a, \varphi, t) \rightarrow \neg Imp(a, \varphi, t)}$

R1. $\dfrac{\ddot{O}bl(a, \varphi, t), a \rightarrow b, \delta \rightarrow \varphi, t \leq t_n : [\ddot{O}bl(z, \psi, t_x) \wedge \delta \rightarrow \psi \wedge a \rightarrow z \wedge t \leq t_x \leq t_n]}{Obl(b, \delta, t_n)}$

---

[1] $L$ is propositional merely for illustration and can be replaced with any formal language with such properties.

$$R2. \frac{\neg I\dddot{m}p(a, \varphi, t), a \to b, \delta \to \varphi, t \leq t_n : [\neg I\dddot{m}p(z, \psi, t_x) \land \delta \to \psi \land a \to z \land t \leq t_x \leq t_n]}{\neg Imp(b, \delta, t_n)}$$

$$R3. \frac{I\dddot{m}p(a, \varphi, t), b \to a, \delta \to \varphi, t \leq t_n : [I\dddot{m}p(z, \psi, t_x) \land \delta \to \psi \land z \to b \to a \land t \leq t_x \leq t_n], \\ [I\dddot{m}p(y, \pi, t_x) \land \delta \to \pi \land b \to y \to a \land t \leq t_x \leq t_n]}{Imp(b, \delta, t_n)}$$

$$R4. \frac{\neg \ddot{O}bl(a, \varphi, t), b \to a, \delta \to \varphi, t \leq t_n : [\neg \ddot{O}bl(z, \psi, t_x) \land \delta \to \psi \land z \to b \to a \land t \leq t_x \leq t_n], \\ [\neg \ddot{O}bl(y, \pi, t_x) \land \delta \to \pi \land b \to y \to a \land t \leq t_x \leq t_n]}{\neg Obl(b, \delta, t_n)}$$

Axioms D1a,b,c,d formalize the standard definitions between deontic modals. Axiom R1 is a conditional version of OB-RM (a similar axiom "CPI" can be found in Olson & Forbus, 2023). It states that if a behavior is stated to be obligatory in a given context, then that agent believes all more general behaviors are obligatory in all more specific contexts. For example, imagine at time $t$ agent $a$ stated, "you must wear a helmet while on a bike": $\ddot{O}bl(a, WearHelmet, OnBike, t)$. Given $[OnBike \land Nighttime] \to OnBike$ and $WearHelmet \to WearHeadProtection$, we can then infer their belief: $Obl(a, WearHeadProtection, OnBike \land Nighttime, t)$. We have then added defeaters to this inference rule for resolving conflicts. This inference is defeated at some future time when the agent has stated a contradictory normative testimony for a more general behavior in an overlapping context. So, obligations are defeated by normative testimony with more general application grounds. We illustrate this in later sections.

Axioms R2-4 formalize inheritance for the other deontic modals. While some deontic accounts take obligation as primitive and leave the other deontic modals implicit, we maintain each and derive corresponding inheritance principles from OB-RM. This allows us to represent normative testimony more naturally without nesting negated statements and is also central for computing defeaters. We explain each of these derived inheritance principles below.

Axiom R3 defines deontic inheritance for prohibitions. It holds that if an agent states that a behavior is impermissible, then that agent believes all more specific behaviors are impermissible as well. Such an inference is defeated when the agent has stated a contradictory normative testimony for an equal or more specific behavior along that entailment path (i.e., either between what was stated and what we are attempting to infer, or under both). Thus, one adds exceptions to prohibitions from below. The inheritance principle underlying this rule is defined and derived from OB-RM below. As a reminder, theorem OB-RM states: $if \vdash p \to q$, $then \vdash Obl(p) \to Obl(q)$.

**Corollary 1** (Inheritance for Prohibitions). $\vdash q \to p$, $then \vdash Imp(p) \to Imp(q)$.

**Proof.** As a reminder, in SDL: $Obl(\neg p) \stackrel{\text{def}}{=} Imp(p)$. Assume, $q \to p$. By Modus Tollens (MT), $\neg p \to \neg q$. Assume $Imp(p)$. By definition, $Obl(\neg p)$. By OB-RM, $Obl(\neg q)$. By definition, $Imp(q)$. Therefore, given $\vdash q \to p$, $\vdash Imp(p) \to Imp(q)$. □

Axioms R2 and R4 then define inheritance for discretionary norms. Together, they hold that a discretionary norm entails that all more general behaviors are non-impermissible, and all more specific behaviors are non-obligatory. This inference is broken out into the two axioms. The former inference (R2) is defeated when a contradictory normative testimony has been stated for an equal or more general behavior. So, like R1, this inference is defeated from above. The latter inference (R4) is defeated when a contradictory normative testimony has been stated for an equal or more specific behavior along that entailment path. So, like R3, R4 gets exceptions added from below. We derive such inferences from OB-RM below.

**Corollary 2** (Inheritance for Discretionary Norms). $\vdash a \to p \to q$, then $\vdash Opt(p) \to \neg Obl(a) \wedge \neg Imp(q)$.

**Proof (1$^{st}$ conjunct).** As a reminder, $Opt(p) \stackrel{def}{=} \neg Obl(p) \wedge \neg Imp(p)$. Assume $a \to p$ and $Opt(p)$. By definition, $\neg Obl(p)$. By MT from OB-RM, $\neg Obl(a)$. Therefore, given $a \to p$ and $Opt(p)$, and thus $\neg Obl(p)$, we can derive $\neg Obl(a)$. □

**Proof (2$^{nd}$ conjunct).** Assume $p \to q$. By MT, $\neg q \to \neg p$. Assume $Opt(p)$. By definition, $\neg Imp(p)$ and thus, $\neg Obl(\neg p)$. By MT from OB-RM, $\neg Obl(\neg q)$. By definition, $\neg Imp(q)$. Therefore, given $p \to q$ and $Opt(p)$, and thus $\neg Imp(p)$, we can derive $\neg Imp(q)$. □

Given these axioms, a *norm structure* of $DDIC$ is a tuple $\langle T, N, D, B, P \rangle$, where $T$ is a set of logical formulae $\in DDIC$ (the background theory), $N$ is a set of deductive rules $= \{D1a, D1b\}$, $D$ is a set of default rules $= \{R1, R2, R3, R4\}$, $B$ is a set of deductive rules $= \{D1c, D1d\}$, and $P$ is a binary relation on $N \cup D \cup B$. Thus, $N$ contains inference for normative testimony, $B$ contains inference for normative beliefs, and $D$ bridges the gap between the two with defeasible inference. Therefore, normative testimony rules are applied before default rules: $\forall (r \in N, d \in D) P(r, d)$. This constraint halts inheritance with future normative testimony. Then, once we infer beliefs from testimony, normative belief inference is applied: $\forall (d \in D, r \in B) P(d, r)$.

## 4. Conflict Resolution Proofs

In this section we demonstrate that DDIC correctly resolves norm conflicts. Along the way, we also reveal that certain norm conflicts from the literature are not truly deontic conflicts. Thus, their "resolution" is merely a product of deontic inheritance. In the proofs below, time flows downwards, illustrating inference as normative testimony is added to the theory. Assume each theory contains the relations between behaviors visualized in Figure 2.

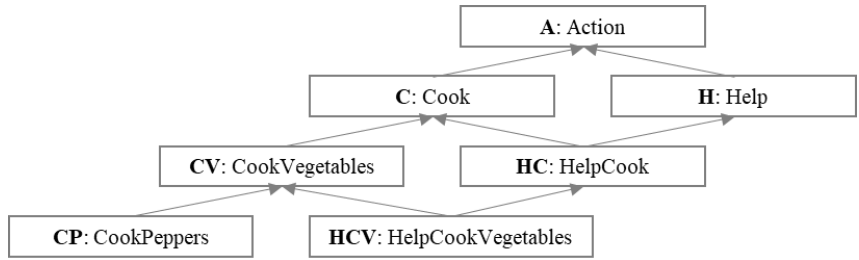

Figure 2. A DAG representing an ontology of generic action types.

## 4.1 Resolving Direct Conflicts

First, consider direct conflicts. As a reminder, two norms are directly conflicting when they share identical application grounds at an intersecting context. For instance, an agent says, "You *must help cook* on Monday", and then "You *cannot help cook* in the morning." Intuitively, their shared application grounds of "helping cook on Monday morning" are impermissible. I.e., the latter norm is preferred (*Lex Posterior*). We prove that DDIC correctly resolves such cases. To save space, we prove only direct conflicts between obligations and prohibitions, but this resolution holds for all direct conflicts.

**Theorem 1** (Given two norms in direct conflict at context $\delta$, the former is completely defeated by the latter at context $\delta$). *If norm structure* $\mathcal{N} = \langle T, N, D, B, P \rangle$, *where* $T = \{\delta \to \varphi, \delta \to \psi, t_1 < t_2 \leq t_n\} \cup ...$
*Case 1a.*: $\{\ddot{O}bl(HC, \varphi, t_1), \dddot{Imp}(HC, \psi, t_2)\}$, *then* $T \vdash Imp(HC, \delta, t_n)$ *and* $T \nvdash Obl(C, \delta, t_n)$;
*Case 1b*: $\{\dddot{Imp}(HC, \psi, t_1), \ddot{O}bl(HC, \varphi, t_2)\}$, *then* $T \vdash Obl(HC, \delta, t_n)$ *and* $T \nvdash Imp(HCV, \delta, t_n)$.

**Proof 1a.** Let norm structure $\mathcal{N} = \langle T, N, D, B, P \rangle$, where $T = \{\delta \to \varphi, \delta \to \psi, t_1 < t_2 \leq t_n\}$.

| | *Statement* | *Reason* |
|---|---|---|
| 1 | $T \vdash \ddot{O}bl(HC, \varphi, t_1)$ | Assumption |
| 2 | $T \vdash \dddot{Imp}(HC, \psi, t_2)$ | Assumption |
| 3 | $T \vdash \neg \ddot{O}bl(HC, \psi, t_2)$ | **MT from D1b & (2)** |
| 4 | $T \vdash Imp(HC, \delta, t_n)$ | R3 from (2), $\delta \to \psi$, & $HC \to HC$ |
| 5 | $T \nvdash Obl(C, \delta, t_n)$ | R1 from (1) defeated by (3), $\delta \to \psi$, & $HC \to HC$ |

Therefore, the derived normative testimony in bold defeats all inheritance from the prior obligation via defeaters in axiom R1 i.e., Lex Posterior. Thus, the agent now believes the acts of helping cook, and thus helping cook vegetables, etc. are impermissible. □

**Proof 1b.** Let norm structure $\langle T, N, D, B, P \rangle$, where $T = \{\delta \to \varphi, \delta \to \psi, t_1 < t_2 \leq t_n\}$.

| | *Statement* | *Reason* |
|---|---|---|
| 1 | $T \vdash \dddot{Imp}(HC, \psi, t_1)$ | Assumption |
| 2 | $T \vdash \ddot{O}bl(HC, \varphi, t_2)$ | Assumption |
| 3 | $T \vdash \neg \dddot{Imp}(HC, \varphi, t_2)$ | **D1b from (2)** |
| 4 | $T \vdash Obl(HC, \delta, t_n)$ | R1 from (2), $\delta \to \varphi$, & $HC \to HC$ |
| 5 | $T \nvdash Imp(HCV, \delta, t_n)$ | R3 from (1) defeated by (3), $\delta \to \varphi$, & $HCV \to HC \to HC$ |

Therefore, the derived normative testimony in bold defeats all downward inference from the prior prohibition via defeaters in axiom 3 i.e., Lex Posterior. □

We illustrate these two cases in Figure 3. Where time flows horizontally, in the top timetable the agent has first stated, "You must help cook." Thus, we can infer that they believe we must cook, help, etc. Then they state, "You cannot help cook." This of course defeats such inferences, and we now infer that they believe helping cook, and all more specific behaviors, are impermissible. With the temporal ordering flipped in the bottom timetable, this defeat occurs in the opposite direction.

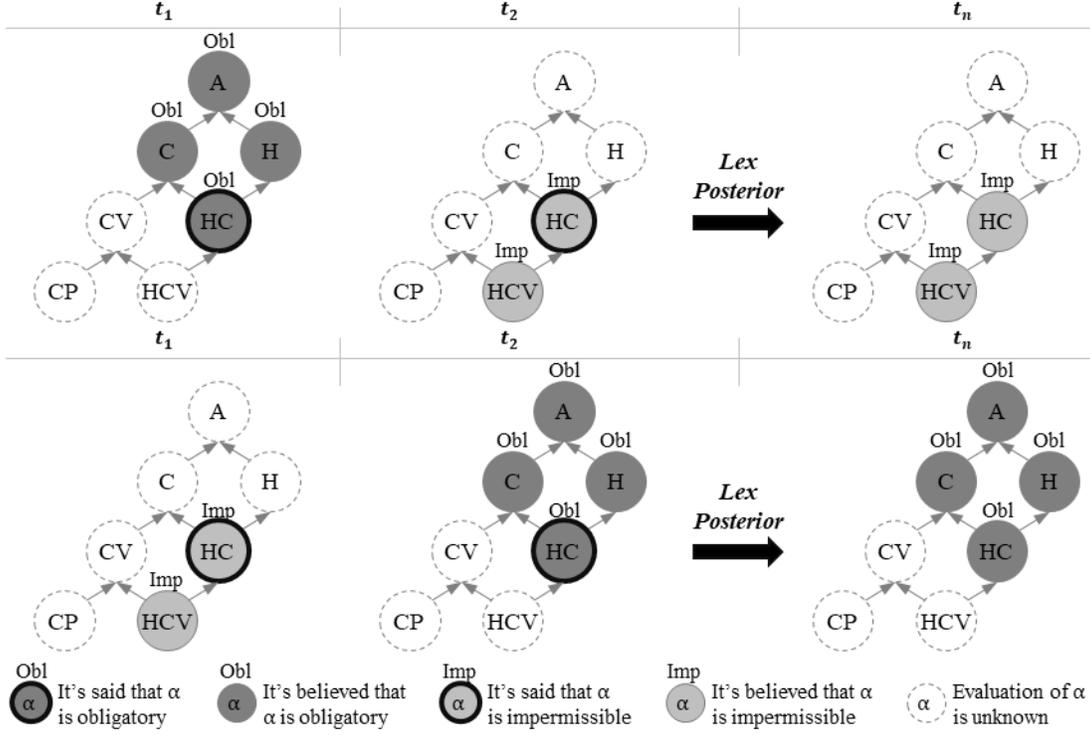

*Figure 3*. DAGs representing direct conflicts between obligations and prohibitions

## 4.2 Resolving Indirect Conflicts

In this section we prove that DDIC correctly resolves all indirect conflicts. As a reminder, two norms are indirectly conflicting when their contexts intersect and the application grounds of one norm entails (yet is not equal to) that of the other norm. That is, one behavior subsumes the other.

### 4.2.1 Indirect Conflicts: Obligations vs Discretionary Norms

We start by considering when the behavior of an obligation is more general than a discretionary norm. For instance, an agent says, "You *must cook* on Monday", and "*Helping cook* in the morning *is optional*." At their shared application grounds "helping cook on Monday morning", the discretionary norm should be preferred (*Lex Specialis*). Contrary to existing analysis, we show that this "resolution" is merely a product of there being no actual conflict, as obligation is not inherited downwards. The first claim does not mean that you must help cook, cook vegetables, etc. on Monday, for any of those will do to satisfy the obligation. Thus, on these shared grounds the more specific discretionary norm prevails. This resolution holds regardless of temporal order.

**Theorem 2** (If an obligation's behavior is more general than that of a discretionary norm at context $\delta$, then their shared application grounds are non-obligatory at $\delta$). *If norm structure* $\mathcal{N} =$ $\langle T, N, D, B, P \rangle$, where $T = \{\ddot{O}bl(C, \varphi, t_x), \ddot{O}pt(HC, \psi, t_y), \delta \to \varphi, \delta \to \psi, t_x \leq t_n, t_y \leq t_n\}$, then $T \vdash \neg Obl(HC, \delta, t_n)$.

**Proof 2.** Let norm structure $\mathcal{N} = \langle T, N, D, B, P \rangle$, where $T = \{\delta \to \varphi, \delta \to \psi, t_x \leq t_n, t_y \leq t_n\}$.

| | Statement | Reason |
|---|---|---|
| 1 | $T \vdash \ddot{O}bl(C, \varphi, t_x)$ | Assumption |
| 2 | $T \vdash \ddot{O}pt(HC, \psi, t_y)$ | Assumption |
| 3 | $T \vdash \neg \ddot{O}bl(HC, \psi, t_y)$ | D1a from (2) |
| 4 | $T \vdash \neg Obl(HC, \delta, t_n)$ | R4 from (3), $\delta \to \psi$, & $HC \to HC$ |

As shown above, these two norms do not conflict under deontic inheritance. However, the more specific discretionary norm does add information, as it explicitly labels all more specific behaviors as non-obligatory and all more general behaviors as non-impermissible. Thus, *Lex Specialis* falls out of deontic inheritance when there is no conflict. □

Next, we consider when a prior discretionary norm governs a more general behavior than a later obligation. The obligation should completely defeat the inferred negated obligations under the discretionary norm. For instance, an agent says, "*Cooking* on Monday is *optional*", and then "You *must help cook* in the morning." Their shared application grounds of "helping cook on Monday morning" should be obligatory i.e., *Lex Posterior*.

**Theorem 3** (If the behavior of a prior discretionary norm is more general than that of an obligation at context $\delta$, then the obligation defeats the discretionary norm only on their shared application grounds at $\delta$). *If norm structure* $\mathcal{N} = \langle T, N, D, B, P \rangle$, *where* $T = \{\ddot{O}pt(C, \varphi, t_1), \ddot{O}bl(HC, \psi, t_2), \delta \to \varphi, \delta \to \psi, t_1 < t_2 \leq t_n\}$, *then* $T \vdash Obl(HC, \delta, t_n), \neg Obl(CV, \delta, t_n)$, *and* $T \nvdash \neg Obl(HCV, \delta, t_n)$.

**Proof 3.** Let norm structure $\mathcal{N} = \langle T, N, D, B, P \rangle$, where $T = \{\delta \to \varphi, \delta \to \psi, t_1 < t_2 \leq t_n\}$.

| | Statement | Reason |
|---|---|---|
| 1 | $T \vdash \ddot{O}pt(C, \varphi, t_1)$ | Assumption |
| 2 | $T \vdash \neg \ddot{O}bl(C, \varphi, t_1)$ | D1a from (1) |
| 3 | **$T \vdash \ddot{O}bl(HC, \psi, t_2)$** | **Assumption** |
| 4 | $T \vdash Obl(HC, \delta, t_n)$ | R1 from (3), $\delta \to \psi$, & $HC \to HC$ |
| 5 | $T \vdash \neg Obl(CV, \delta, t_n)$ | R4 from (2), $\delta \to \varphi$, & $CV \to C$ |
| 6 | $T \nvdash \neg Obl(HCV, \delta, t_n)$ | R4 from (2) defeated by (3), $\delta \to \psi$, & $HCV \to HC \to C$ |

As shown above, these two norms conflict at $\delta$ between behaviors $HC$ and $C$. Therefore, when trying to infer downwards from the prior discretionary norm via R4, the more specific obligation stated later defeats it along the path they share[2]. However, the downward inference from the discretionary norm that does not share a path with the later obligation remains. □

Next, we consider the reverse order, when a more specific obligation comes prior to a discretionary norm. The obligation should be completely defeated. For instance, an agent says, "You *must help cook* on Monday", and then "*Cooking* in the morning is *optional*." Their shared application grounds of "helping cook on Monday morning" should be non-obligatory i.e., *Lex Posterior*.

---

[2] Note that shared inferences upwards from the prior discretionary norm to "non-impermissible" are not defeated. However, they are *strengthened* by the obligation i.e., from "this is permissible" to "this is obligatory."

**Theorem 4** (If the behavior of a prior obligation is more specific behavior than that of a later discretionary norm at context $\delta$, then the discretionary norm completely defeats the obligation at $\delta$). *If norm structure $\mathcal{N} = \langle T, N, D, B, P \rangle$, where $T = \{\ddot{O}bl(HC, \psi, t_1), \dddot{Opt}(C, \varphi, t_2), \delta \to \varphi, \delta \to \psi, t_1 < t_2 \leq t_n\}$, then $T \vdash \neg Obl(HC, \delta, t_n)$ and $T \nvdash Obl(H, \delta, t_n)$.*

**Proof 4.** Let norm structure $\mathcal{N} = \langle T, N, D, B, P \rangle$, where $T = \{\delta \to \varphi, \delta \to \psi, t_1 < t_2 \leq t_n\}$.

    *Statement*                 *Reason*
1.  $T \vdash \ddot{O}bl(HC, \psi, t_1)$      Assumption
2.  $T \vdash \dddot{Opt}(C, \varphi, t_2)$       Assumption
3.  **$T \vdash \neg \ddot{O}bl(C, \varphi, t_2)$**    **D1a from (2)**
4.  $T \vdash \neg Obl(HC, \delta, t_n)$    R4 from (3), $\delta \to \varphi$, & $HC \to C$
5.  $T \nvdash Obl(H, \delta, t_n)$        R1 from (1) defeated by (3), $\delta \to \varphi$, & $HC \to C$

Therefore, the later more general discretionary norm completely defeats all inferences made from the prior obligation via justifications in R1. □

Next, we examine indirect conflicts between prohibitions and discretionary norms.

### 4.2.2 Indirect Conflicts: Prohibitions vs Discretionary Norms

We start by considering when the behavior of a prohibition is more specific behavior than that of a discretionary norm. Regardless of order, the prohibition should be preferred. For instance, an agent says, "You *cannot cook vegetables* on Monday", and "*Cooking* in the morning is *optional*." Their shared application grounds of "cooking vegetables on Monday morning" is still impermissible i.e., *Lex Specialis*. We again show that this merely results from there being no true conflict.

**Theorem 5** (If the behavior of a prohibition is more specific than that of a discretionary norm at context $\delta$, then their shared application grounds are impermissible at $\delta$). *If norm structure $\mathcal{N} = \langle T, N, D, B, P \rangle$, where $T = \{\dddot{Imp}(CV, \varphi, t_x), \dddot{Opt}(C, \psi, t_y), \delta \to \varphi, \delta \to \psi, t_x \leq t_n, t_y \leq t_n\}$, then $T \vdash Imp(CV, \delta, t_n)$.*

**Proof 5.** Let norm structure $\mathcal{N} = \langle T, N, D, B, P \rangle$, where $T = \{\delta \to \varphi, \delta \to \psi, t_x \leq t_n, t_y \leq t_n\}$.

    *Statement*                 *Reason*
1.  $T \vdash \dddot{Imp}(CV, \varphi, t_x)$    Assumption
2.  $T \vdash \dddot{Opt}(C, \psi, t_y)$     Assumption
3.  $T \vdash \neg \dddot{Imp}(C, \psi, t_y)$   D1a from (2)
4.  $T \vdash Imp(CV, \delta, t_n)$      R3 from (1), $\delta \to \varphi$, & $CV \to CV$

As seen above, the two norms do not conflict under deontic inheritance. Thus, as the prohibition is more specific and inherits downwards, their shared application grounds are impermissible, regardless of order i.e., *Lex Specialis*. □

Next, we consider cases when a prohibition is more general than a discretionary norm. We start with the case when a prohibition is stated first. The strategy here should be to prioritize the discretionary norm. For instance, an agent says, "You *cannot cook* on Monday", and then "*Helping cook* in the morning is *optional*". The discretionary norm adds exceptions at their shared grounds "helping cook on Monday morning", defeating the impermissibility i.e., *Lex Posterior*.

**Theorem 6** (If the behavior of a prior prohibition is more general than that of a later discretionary norm at context $\delta$, then the discretionary norm defeats the prohibition only at their shared application grounds at $\delta$). *If norm structure $\mathcal{N} = \langle T, N, D, B, P \rangle$, where $T =$* $\{\dddot{Imp}(C,\varphi,t_1), \ddot{Opt}(HC,\psi,t_2), \delta \to \varphi, \delta \to \psi, t_1 < t_2 \leq t_n\}$,
then $T \vdash \neg Imp(HC,\delta,t_n), Imp(CV,\delta,t_n)$ and $T \nvdash Imp(HCV,\delta,t_n)$.

**Proof 6.** Let norm structure $\mathcal{N} = \langle T, N, D, B, P \rangle$, where $T = \{\delta \to \varphi, \delta \to \psi, t_1 < t_2 \leq t_n\}$.

| | Statement | Reason |
|---|---|---|
| 1 | $T \vdash \dddot{Imp}(C,\varphi,t_1)$ | Assumption |
| 2 | $T \vdash \ddot{Opt}(HC,\psi,t_2)$ | Assumption |
| **3** | $\mathbf{T \vdash \neg \dddot{Imp}(HC,\psi,t_2)}$ | **D1a from (2)** |
| 4 | $T \vdash \neg Imp(HC,\delta,t_n)$ | R2 from (3), $\delta \to \psi$, & $HC \to HC$ |
| 5 | $T \vdash Imp(CV,\delta,t_n)$ | R3 from (1), $\delta \to \varphi$, & $CV \to C$ |
| 6 | $T \nvdash Imp(HCV,\delta,t_n)$ | R3 from (1) defeated by (3), $\delta \to \psi$, & $HCV \to HC \to C$ |

As shown above, the inheritance from the previous prohibition is correctly halted along the path shared by the discretionary norm via defeaters in R3 i.e. *Lex Posterior*. However, inheritance at behaviors not along this path remains. That is, the discretionary norm adds exceptions. □

Next, we consider when the discretionary norm is stated first. The strategy here should be to prioritize the prohibition. For instance, an agent says, "*Helping cook* on Monday is optional", and then "You *cannot cook* in the morning." Their shared application grounds of "helping cook on Monday morning" are impermissible i.e., *Lex Posterior*.

**Theorem 7** (If the behavior of a prior discretionary norm is more specific than that of a later prohibition at context $\delta$, then the prohibition completely defeats upwards inference from the discretionary norm at $\delta$). *If norm structure* $\mathcal{N} = \langle T, N, D, B, P \rangle$, *where* $T = \{\ddot{Opt}(HC,\psi,t_1), \dddot{Imp}(C,\varphi,t_2), \delta \to \varphi, \delta \to \psi, t_1 < t_2 \leq t_n\}$, *then* $T \vdash Imp(HC,\delta,t_n)$ *and* $T \nvdash \neg Imp(H,\delta,t_n)$.

**Proof 7.** Let norm structure $\mathcal{N} = \langle T, N, D, B, P \rangle$, where $T = \{\delta \to \varphi, \delta \to \psi, t_1 < t_2 \leq t_n\}$.

| | Statement | Reason |
|---|---|---|
| 1 | $T \vdash \ddot{Opt}(HC,\psi,t_1)$ | Assumption |
| 2 | $T \vdash \neg \dddot{Imp}(HC,\psi,t_1)$ | D1a from (1) |
| **3** | $\mathbf{T \vdash \dddot{Imp}(C,\varphi,t_2)}$ | **Assumption** |
| 4 | $T \vdash Imp(HC,\delta,t_n)$ | R3 from (3), $\delta \to \varphi$, & $HC \to C$ |
| 5 | $T \nvdash \neg Imp(H,\delta,t_n)$ | R2 from (2) defeated by (3), $\delta \to \varphi$, & $HC \to C$ |

As shown above, the upwards inheritance from the previously stated discretionary norm is completely defeated by the later prohibition via defeaters in R2 i.e., *Lex Posterior*. □

*4.2.3 Indirect Conflicts: Obligations vs Prohibitions*

Next, we examine indirect conflicts between obligations and prohibitions. We start with when an obligation is more general. We show that there is no conflict under deontic inheritance in such cases and thus why the prohibition is preferred regardless of order. For instance, an agent says, "You *must cook* on Monday", and "You *cannot cook vegetables* in the morning." Their shared application grounds of "cooking vegetables on Monday morning" are impermissible i.e., *Lex Specialis*.

**Theorem 8** (If the behavior of an obligation is more general than that of a prohibition at context $\delta$, then their shared application grounds are impermissible at $\delta$). *If norm structure* $\mathcal{N} =$ $\langle T, N, D, B, P \rangle$, where $T = \{\dddot{Obl}(C, \psi, t_x), \dddot{Imp}(CV, \varphi, t_y), \delta \to \varphi, \delta \to \psi, t_x \leq t_n, t_y \leq t_n\}$, then $T \vdash Imp(CV, \delta, t_n)$.

**Proof 8.** Let norm structure $\mathcal{N} = \langle T, N, D, B, P \rangle$, where $T = \{\delta \to \varphi, \delta \to \psi, t_x \leq t_n, t_y \leq t_n\}$.

| | Statement | Reason |
|---|---|---|
| 1 | $T \vdash \dddot{Obl}(C, \psi, t_x)$ | Assumption |
| 2 | $T \vdash \dddot{Imp}(CV, \varphi, t_y)$ | Assumption |
| 3 | $T \vdash Imp(CV, \delta, t_n)$ | R3 from (2), $\delta \to \varphi$, & $CV \to CV$ |

As shown above, the two norms never conflict under deontic inheritance. Trivially, the more general obligation labels all more general behaviors as obligatory and the more specific prohibition labels all more specific behaviors as impermissible. Therefore, their shared application grounds are impermissible i.e., *Lex Specialis*. □

Next, we consider when a prohibition is more general than an obligation. When it is stated first, the later obligation should be preferred. For instance, an agent says, "You *cannot cook vegetables* on Monday", and then "You *must help cook vegetables* in the morning." Their shared application grounds of "helping cook vegetables on Monday morning" are obligatory i.e., *Lex Posterior*.

**Theorem 9** (If the behavior of a prior prohibition is more general than that of a later obligation at context $\delta$, then the obligation defeats the prohibition only at their shared application grounds at $\delta$). *If norm structure $\mathcal{N} = \langle T, N, D, B, P \rangle$, where $T = \{\dddot{Imp}(CV, \varphi, t_1), \dddot{Obl}(HCV, \psi, t_2), \delta \to \varphi, \delta \to \psi, t_1 < t_2 \leq t_n\}$, then $T \vdash Obl(HCV, \delta, t_n), Imp(CP, \delta, t_n)$ and $T \nvdash Imp(HCV, \delta, t_n)$.*

**Proof 9.** Let norm structure $\mathcal{N} = \langle T, N, D, B, P \rangle$, where $T = \{\delta \to \varphi, \delta \to \psi, t_1 < t_2 \leq t_n\}$.

| | Statement | Reason |
|---|---|---|
| 1 | $T \vdash \dddot{Imp}(CV, \varphi, t_1)$ | Assumption |
| 2 | $T \vdash \dddot{Obl}(HCV, \psi, t_2)$ | Assumption |
| **3** | $\boldsymbol{T \vdash \neg \dddot{Imp}(HCV, \psi, t_2)}$ | **D1b from (2)** |
| 4 | $T \vdash Obl(HCV, \delta, t_n)$ | R1 from (2), $\delta \to \psi$, & $HCV \to HCV$ |
| 5 | $T \vdash Imp(CP, \delta, t_n)$ | R3 from (1), $\delta \to \varphi$, & $CP \to CV$ |
| 6 | $T \nvdash Imp(HCV, \delta, t_n)$ | R3 from (1) defeated by (3), $\delta \to \psi$, & $HCV \to HCV \to CV$ |

As shown above, the boldened formulae derived at time $t_2$ defeats the path of downward inheritance from the prior prohibition. However, the inheritance at $CP = CookPeppers$ has not been defeated. Thus, the later obligation adds exceptions i.e., *Lex Posterior*. □

Next, we consider the reverse order, when a more specific obligation comes prior to a more general prohibition. The prohibition should completely defeat the obligation. For instance, an agent says, "You *must help cook vegetables* on Monday", and then "You *cannot cook vegetables* in the morning." Their shared application grounds of "helping cook vegetables on Monday morning" should be impermissible i.e., *Lex Posterior*.

**Theorem 10** (If the behavior of a prior obligation is more specific than that of a prohibition at context $\delta$, then the prohibition completely defeats the obligation at $\delta$). *If norm structure $\mathcal{N} = \langle T, N, D, B, P \rangle$, where $T = \{\dddot{Obl}(HCV, \psi, t_1), \dddot{Imp}(CV, \varphi, t_2), \delta \to \varphi, \delta \to \psi, t_1 < t_2 \leq t_n\}$, then $T \vdash Imp(HCV, \delta, t_n)$ and $T \nvdash Obl(HC, \delta, t_n)$.*

**Proof 10.** Let norm structure $\mathcal{N} = \langle T, N, D, B, P \rangle$, where $T = \{\delta \to \varphi, \delta \to \psi, t_1 < t_2 \leq t_n\}$.

| | Statement | Reason |
|---|---|---|
| 1 | $T \vdash \ddot{O}bl(HCV, \psi, t_1)$ | Assumption |
| 2 | $T \vdash \ddot{Imp}(CV, \varphi, t_2)$ | Assumption |
| 3 | $T \vdash \neg \ddot{O}bl(CV, \varphi, t_2)$ | **MT from D1b & (2)** |
| 4 | $T \vdash Imp(HCV, \delta, t_n)$ | R3 from (2), $\delta \rightarrow \varphi$, & $HCV \rightarrow CV$ |
| 5 | $T \nvdash Obl(HC, \delta, t_n)$ | R1 from (1) defeated by (3), $\delta \rightarrow \varphi$, & $HCV \rightarrow CV$ |

As shown above, the boldened formula derived at time $t_2$ defeats all upwards inheritance from the prior obligation i.e., *Lex Posterior*. □

In Figure 4 below, we further illustrate the order-dependency of these two cases. In the top timetable, when a more specific obligation comes after a prohibition, it defeats certain paths of inheritance. In our example, the agent stated, "do not cook vegetables" and then, "you must help cook vegetables." They have overridden the prohibition only at a specific type of cooking vegetables (helping). However, with the reverse order illustrated in the bottom table, we take the speaker to be completely overriding the more specific prior obligation.

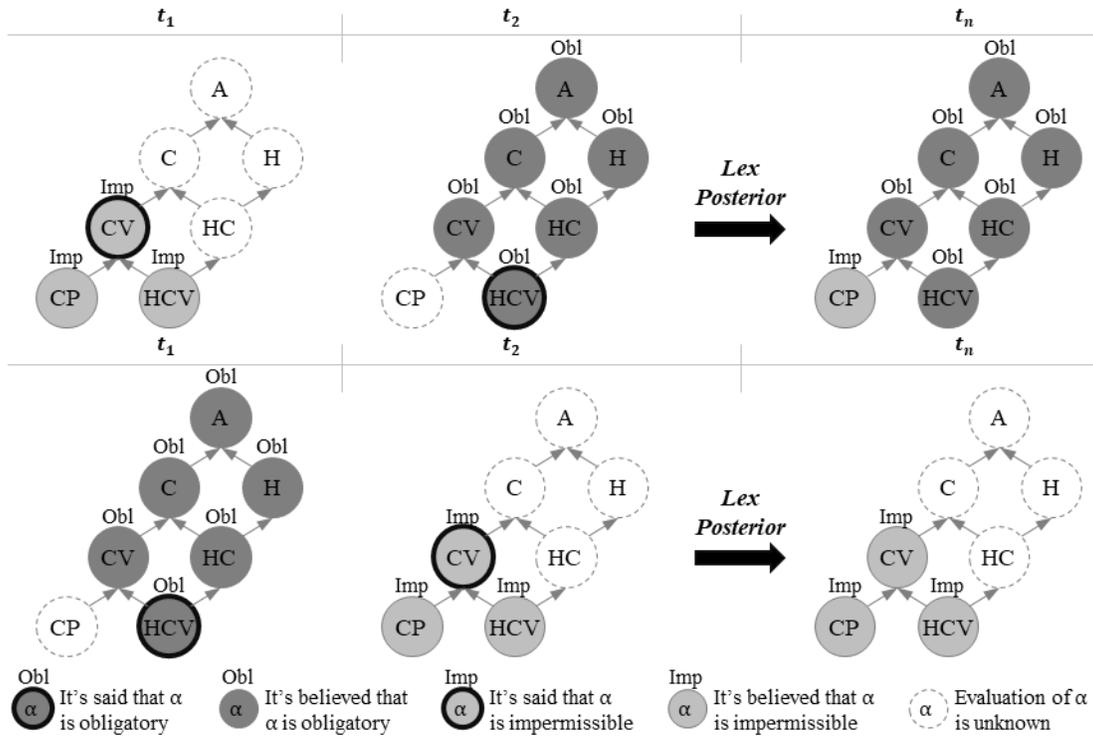

*Figure 4.* DAGs detailing the order-dependency of resolution between a general prohibition and a more specific obligation.

### 4.3 Resolving Intersecting Conflicts

In this section we consider intersecting conflicts within DDIC. Recall that two norms conflict at an intersection when their application grounds share a proper subset, and their deontic statuses are inconsistent. However, we show that in all such cases, there is no true deontic conflict under inheritance and how the corresponding resolution strategy falls out of this fact.

*4.3.1 Intersecting Conflicts: Obligations vs Discretionary Norms*

We start by considering intersections between an obligation and a discretionary norm. The resolution strategy here should be to prefer the discretionary norm, regardless of order. For instance, an agent says, "You *must cook vegetables* on Monday", and "*Helping cook* in the morning is *optional*." Their shared application grounds of "helping cook vegetables on Monday morning" are non-obligatory.

**Theorem 11** (If the behavior of an obligation and a discretionary norm intersect at $S$ in context $\delta$, then $S$ is non-obligatory at $\delta$). *If norm structure $\mathcal{N} = \langle T, N, D, B, P \rangle$, where $T = \{\ddot{O}bl(CV, \varphi, t_x), \ddot{O}pt(HC, \psi, t_y), \delta \to \varphi, \delta \to \psi, t_x \leq t_n, t_y \leq t_n\}$, then $T \vdash \neg Obl(HCV, \delta, t_n)$.*

**Proof 11.** Let norm structure $\mathcal{N} = \langle T, N, D, B, P \rangle$, where $T = \{\delta \to \varphi, \delta \to \psi, t_x \leq t_n, t_y \leq t_n\}$.

| | Statement | Reason |
|---|---|---|
| 1 | $T \vdash \ddot{O}bl(CV, \varphi, t_x)$ | Assumption |
| 2 | $T \vdash \ddot{O}pt(HC, \psi, t_y)$ | Assumption |
| 3 | $T \vdash \neg \ddot{O}bl(HC, \psi, t_y)$ | D1a from (2) |
| 4 | $T \vdash \neg Obl(HCV, \delta, t_n)$ | R4 from (3), $\delta \to \psi$, & $HCV \to HC$ |

As shown above, because obligations do not inherit downwards, the two norms never conflict. However, they do complement each other. The discretionary norm labels more specific behaviors as non-obligatory, and the obligation labels more general behaviors as obligatory. Therefore, inference from the discretionary norm is prioritized at their shared application grounds. □

*4.3.2 Intersecting Conflicts: Prohibitions vs All Norm Types*

Next, we consider intersections between prohibitions and all other norm types. The resolution strategy here should be to prefer the prohibition, regardless of order. For instance, an agent says, "You *cannot cook vegetables* on Monday", and "*Helping cook* in the morning is *optional*." Their shared applications grounds "helping cook vegetables on Monday morning" are impermissible. We prove this only for intersections with discretionary norms, but this holds with obligations as well.

**Theorem 12** (If the behavior of a prohibition and a discretionary norm intersect at S in context $\delta$, then S is impermissible at $\delta$). *If norm structure $\mathcal{N} = \langle T, N, D, B, P \rangle$, where $T = \{I\ddot{m}p(CV, \varphi, t_x), \ddot{O}pt(HC, \psi, t_y), \delta \to \varphi, \delta \to \psi, t_x \leq t_n, t_y \leq t_n\}$, then $T \vdash Imp(HCV, \delta, t_n)$.*

**Proof 12.** Let norm structure $\mathcal{N} = \langle T, N, D, B, P \rangle$, where $T = \{\delta \to \varphi, \delta \to \psi, t_x \leq t_n, t_y \leq t_n\}$.

| | Statement | Reason |
|---|---|---|
| 1 | $T \vdash I\ddot{m}p(CV, \varphi, t_x)$ | Assumption |
| 2 | $T \vdash \ddot{O}pt(HC, \psi, t_y)$ | Assumption |
| 3 | $T \vdash Imp(HCV, \delta, t_n)$ | R3 from (1), $\delta \to \varphi$, & $HCV \to CV$ |

Again, these two intersecting norms do not conflict under deontic inheritance (though they do complement each other). Thus, because prohibitions inherit downwards via R3, their shared application grounds are impermissible. Without loss of generality, the same is true for intersections between prohibitions and obligations. □

## 5. Related Work

Automatically detecting and resolving norm conflicts is part of active research on normative multi-agent systems (NMAS) (Santos et al., 2017). For instance, Da Silva and Zahn (2014) examined contextual relations and inheritance rules to evaluate potential conflicts between two norms. They then defined a novel algorithm for checking for norm conflicts and rewriting the norms to resolve them. However, it is quite difficult to maintain a long chain of norm edits and their dependencies over time. Our defeasible reasoning approach yields a much simpler norm update mechanism.

An approach similarly grounded in deontic logic is that of Cholvy and Cuppens (2000). Their logic FUSION resolves conflicts via preference orderings amongst norms. However, they did not explicitly consider conditional norms (though they do briefly discuss it for future work). We provide an analysis of defeasibility considering entailments between both behaviors and contexts.

Horty (1994) similarly explored default logic for resolving deontic conflicts. While we draw inspiration from Horty here, he only considered defeasibility in the context of a norm, while we consider the behavior as well. Thus, our approach can detect and resolve indirect and intersecting conflicts. Furthermore, being more concerned with legal theory, he considered norms as an order-independent set. However, here we are concerned with resolving the ever-changing normative beliefs of an agent, and the order in which that agent states their norms is fundamental.

## 6. Conclusion and Future Work

We have presented a novel defeasible deontic calculus with inheritance (DDIC) and proven that it resolves norm conflicts. We have demonstrated that the resolution strategy Lex Posterior falls out of two norms being in contradiction under deontic inheritance. We then revealed the strategy of Lex Specialis as a red herring. Insofar as deontic inheritance is viewed as a plausible inference, preference for a more specific norm merely results from deontic inheritance without conflict. These findings are theoretically interesting as they result in an axiomatization for norm conflict resolution that is rooted in standard deontic logic. This axiomatization, DDIC, is also practically useful, as it resolves norm conflicts without having to maintain edits like other approaches.

We plan to explore three avenues of future work. First, while we only considered conflicts within a single normative theory here, we plan to explore merging across theories, particularly across diverse individuals and institutions. Second, we plan to consider speaker reliability and ambiguity (as in Olson and Forbus, 2021) for norm conflict resolution. We have taken a speaker's testimony to directly imply their belief, but of course speakers can lie, and natural language is often ambiguous. Considering these features will allow more complex and abstract norms to be handled. Lastly, the work here has been proof-theoretic, but we are also implementing and testing this theory in the Companion cognitive architecture (Forbus & Hinrichs, 2017). This system will act in accordance with users' norms, while users can continually update their norms via natural language.

## Acknowledgements

This research was sponsored by the US Air Force Office of Scientific Research under award number FA95550-20-1- 0091. Taylor Olson was supported by an IBM Fellowship. We thank Constantine Nakos, Mukundan Kuthalam, and anonymous reviewers for their feedback on earlier drafts.